\documentclass[runningheads]{llncs}

\usepackage{eccv}

\usepackage{eccvabbrv}

\usepackage{graphicx}
\usepackage{booktabs}
\usepackage{tikz}
\usepackage{footnote}
\usepackage[accsupp]{axessibility}  

\usepackage{hyperref}

\usepackage{orcidlink}

\usepackage{graphicx}
\usepackage{hyperref}
\usepackage{graphicx}
\usepackage{float}
\usepackage{xcolor}
\usepackage[many]{tcolorbox}
\tcbuselibrary{listings}
\usepackage{listings}
\usepackage{graphicx}
\usepackage{booktabs}
\usepackage{adjustbox}
\usepackage{algorithm}
\usepackage{algpseudocode}
\usepackage{amsmath}
\usepackage{float}
\usepackage{tikz}
\usepackage{tabularx}
\usepackage{caption}
\usepackage{subcaption}

\begin{document}
	
	
	\title{Efficient NeRF Optimization - Not All Samples Remain Equally Hard} 

	\author{ Juuso Korhonen\inst{1} \and
		Goutham Rangu\inst{1} \and
		Hamed R. Tavakoli\inst{1} \orcidlink{0000-0002-9466-9148} \and
		Juho Kannala\inst{2,3}}
	
	\authorrunning{J. Korhonen, et al.}
	
	\institute{Nokia Technologies, Finland \\
		\email{\{juuso.korhonen, goutham.rangu, hamed.rezazadegan\_tavakoli\}@nokia.com}\\
		\and
		Aalto University, Finland \\
		\and University of Oulu, Finland\\
		\email{juho.kannala@aalto.fi} \\
	}
	
	\maketitle

	\begin{abstract}
		We propose an application of online hard sample mining for efficient training of Neural Radiance Fields (NeRF). NeRF models produce state-of-the-art quality for many 3D reconstruction and rendering tasks but require substantial computational resources. The encoding of the scene information within the NeRF network parameters necessitates stochastic sampling. We observe that during the training, a major part of the compute time and memory usage is spent on processing already learnt samples, which no longer affect the model update significantly. We identify the backward pass on the stochastic samples as the computational bottleneck during the optimization. We thus perform the first forward pass in inference mode as a relatively low-cost search for hard samples. This is followed by building the computational graph and updating the NeRF network parameters using only the hard samples. To demonstrate the effectiveness of the proposed approach, we apply our method to Instant-NGP, resulting in significant improvements of the view-synthesis quality over the baseline (1 dB improvement on average per training time, or 2x speedup to reach the same PSNR level) along with $\sim$40\% memory savings coming from using only the hard samples to build the computational graph. As our method only interfaces with the network module, we expect it to be widely applicable.
		\keywords{Neural radiance fields \and Importance sampling \and Efficient optimization}
	\end{abstract}

	\section{Introduction}
	\label{sec:intro}
	Neural radiance field (NeRF) modeling, first introduced in \cite{mildenhall2020nerf}, has attracted significant attention from the research community leading to extensive efforts in its further development and application. NeRF models represent state-of-the-art solutions for photorealistic novel view-synthesis with view-dependent effects. Given the impressive results, the volumetric rendering pipeline at the core of NeRFs has been applied to many tasks in 3D reconstruction and rendering \cite{martin2021nerf, yu2021plenoctrees, barron2022mipnerf, tancik2022block, li2023instant3d}.

	However, the state-of-the-art NeRF models for different 3D reconstruction metrics like view-synthesis quality \cite{barron2022mipnerf}, dynamic scene-\cite{song2023nerfplayer, park2021hypernerf, fridovichkeil2023kplanes}, and surface reconstruction \cite{li2023neuralangelo}, require long training times and substantial memory usage, largely due to the stochastic sampling required by the implicit nature of the NeRF models. The recent explicit derivatives of NeRF \cite{yu2021plenoxels, 3dgs} combat this, achieving excellent quality and training/rendering speed, but at a cost of massively increased memory usage. The Instant-NGP \cite{mueller2022instant}, which uses occupancy grids and multi-resolution hash-encoding paired with small and efficient neural networks, still has one of the best trade-offs for training speed, memory usage and achieved quality.
	
	One aspect that has received limited attention is hard sample mining during the NeRF optimization process.  The standard random selection of pixels for the NeRF volume rendering process floods the training with easy samples, neglecting whether specific 3D locations (coming from specific rays) have already been sufficiently modeled, and leads to unnecessary and costly neural network processing. The relevant guided ray sampling methods \cite{barron2022mipnerf, mueller2022instant} focus the samples near the scene surfaces, but are still agnostic to the question if these samples are already modeled well enough. 
	
	To improve the learning that we achieve per sample processing, we propose a hard sample mining method for the NeRF optimization. Leveraging the computational characteristics of a neural network optimization, we perform a forward pass in an inference mode to conduct a low-cost search for hard point samples. The hardness of a point sample is determined based on the pixel loss that gets backpropagated over the volume rendering to the network outputs for the point sample. The proposed algorithm automatically adjusts the hard sample batch size to achieve on average the same convergence per iteration as the original batch. Only the hard sample batch undergoes the construction of a computational graph, and model parameters are updated accordingly with reduced training iteration time and memory requirements. We show that computational resources are more effectively utilized by focusing on the hard samples of the scene during NeRF optimization. We apply the proposed hard sample mining method to the optimization of the Instant-NGP model \cite{mueller2022instant} to demonstrate the benefits of the proposed method. In summary, our contributions are:

	\begin{itemize}
		\item Introduction of a hyperparameter-free method to mine for hard point samples in an online manner during each training iteration.
		\item Reduction in the training iteration time and memory requirements by creating and backpropagating the network computational graph only for the hard point sample minibatch.
		\item Significant improvement to view-synthesis quality with reduced training time and memory budget.
	\end{itemize}
	
	We attribute the effectiveness of our method to excluding the expensive network backpropagation for the samples that do not significantly contribute to the model updates. These samples originate not only from well-modeled locations but, as our investigation reveals, also include occluded and empty space samples that persist in the batch following the pruning of the ray sampling. As our method only interfaces with the NeRF network module, we expect its applicability being widespread.
	
	\section{Related Work}
	
	\subsubsection{Neural radiance fields} Despite the wide variety of NeRF methods and applications, most of them share a common volumetric rendering pipeline: (1) casting a ray through an image pixel, (2) ray sample for points along the ray, (3) evaluate the point samples with a neural network producing density $\sigma$ and color $c$ estimates per point, (4) accumulate the point attributes onto a pixel color using volume rendering:
	
	\begin{equation}
		\label{volrend_eq}
		C(r) = \sum_{i=1}^{N} w_i c_i, \text{ where } w_i = T_i \alpha_i, T_i = \prod_{j=1}^{i-1} (1 - \alpha_j), \alpha_i = 1 - \exp(-\sigma_i \delta_i)
	\end{equation}
	
	\noindent where $w$ is the point sample weight for the pixel color of the ray $C(r)$, $T$ is the transmittance at the point sample, and $\delta$ is the step size until next point sample on the ray. Comprehensive survey of how this pipeline has been tuned and applied to different 3D reconstruction tasks is given in \cite{gao2023nerfsurvey}, but by and large it has remained to share the same structure.
	
	The implicit and continuous way the scene information is encoded within the NeRF network parameters necessitates stochastic sampling. The computational bottleneck of both the NeRF reconstruction and the rendering comes from the massive amount of network sample processing required by the stochastic sampling to render an image.
	
	\subsubsection{Guided ray sampling}  As sampling densely along the ray would be computationally too expensive, guided ray sampling strategies have been pivotal in speeding up the volumetric rendering process. The main objective for the guided ray sampling methods has been pruning the low weight ($w$) samples, since they have a negligible effect on the resulting pixel color.
	
	Two most prominent ray sampling methods, grid sampling \cite{mueller2022instant} and proposal sampling \cite{barron2022mipnerf} aim to distill the density knowledge of the main NeRF network onto a faster-to-query data structure. Grid sampling utilizes a separate occupancy grid, where cell values get updated in a stochastic way with density estimates from the further optimized main network. Proposal sampling utilizes a sequence of lightweight proposal networks, which learn to bound the main network predicted weights.
	
	Both approaches have improved the speed of the rendering process drastically by reducing the number of evaluations required of the main NeRF network. However, these methods achieve the increase in the rendering speed with an increase in the model size and with the extra storage requirement for the occupancy grid or the proposal networks.
	
	There are some limitations to the low-weight sample pruning of these methods. The proposal samplers do concentrate more samples on the first surface site, but some samples might still end up in empty space due to the iterative PDF resampling starting from uniform sampling. Also, by default, grid sampling does not prune occluded samples, since that requires rendering the transmittance. We however point the reader to a subsequent approach by Li et al. \cite{li2023nerfacc} where the transmittance is first rendered in inference mode, and only the high-weight samples are resampled for the training iteration. This approach shares similarities to our two-forward-one-backward approach, but does not take into account the loss, i.e. if we already handle these samples well according to the pixel loss they contribute to.
	
	It is important to note that neither of the ray sampling methods in \cite{barron2022mipnerf, mueller2022instant} belong to the same computational graph as the main network. The offline distillation of the grid sampling does not build a computational graph, and the online distillation of the proposal sampling uses a detached versions of the main network predicted weights. As our method acts on the main network computational graph, these ray sampling methods are complementary and are left unaffected.
	
	\subsubsection{Guided pixel sampling} Less prominent guided pixel sampling approaches aim to be selective for which pixels do we cast rays and apply the optimization in the first place. \cite{egranerf} and \cite{saliency} use pre-trained networks to acquire initial pixel sampling images. \cite{sun2023efficient} increases sampling probability for pixels with high neighboring view color variance and neighboring pixel depth variance. Relevant to our method, previous implementations of hard sample mining are found in \cite{otonari2022nonuniform} and \cite{wang2022r2l}, where historical pixel losses are used to derive the pixel sampling weights. A common problem to these approaches is that the historical loss might not accurately reflect the current performance due to carried out model updates. The guided pixel sampling approaches use hyper-parameterized schemes to ensure sufficient coverage and to modulate the historical losses. Most of the existing NeRF methods, even those representing state-of-the-art with respect to the optimization speed like \cite{mueller2022instant, yu2021plenoxels, 3dgs}, have remained using random sampling as their pixel sampling strategy.
	
	\subsubsection{Other approaches increasing reconstruction \& rendering speed}
	
	Other major approach to increase the speed of the volumetric rendering pipeline are the parametrized input-encoding strategies \cite{chen2022tensorf, fridovichkeil2023kplanes, mueller2022instant, fridovich2022plenoxels}  that allow to use smaller and faster neural networks or even forgo the neural network processing completely. These approaches however come with a significant increase in the model size due to the storage of the input-encoding parameters such as hash-tables or feature grids.
	
	There exist also a category of models which aim to speed up specifically the inference by baking the NeRF model into some faster-to-query data structure \cite{hedman2021bakingsnerg, yu2021plenoctrees, garbin2021fastnerf, reiser2021kilonerf, reiser2023merf}. The real-time rendering however comes at a cost of training time, achieved quality and massive increase in the model size.

	\section{Hard Sample Mining for NeRF Optimization}
	\label{methods}
	
\begin{figure}[t]
		\centering
		\includegraphics[width=0.8\linewidth]{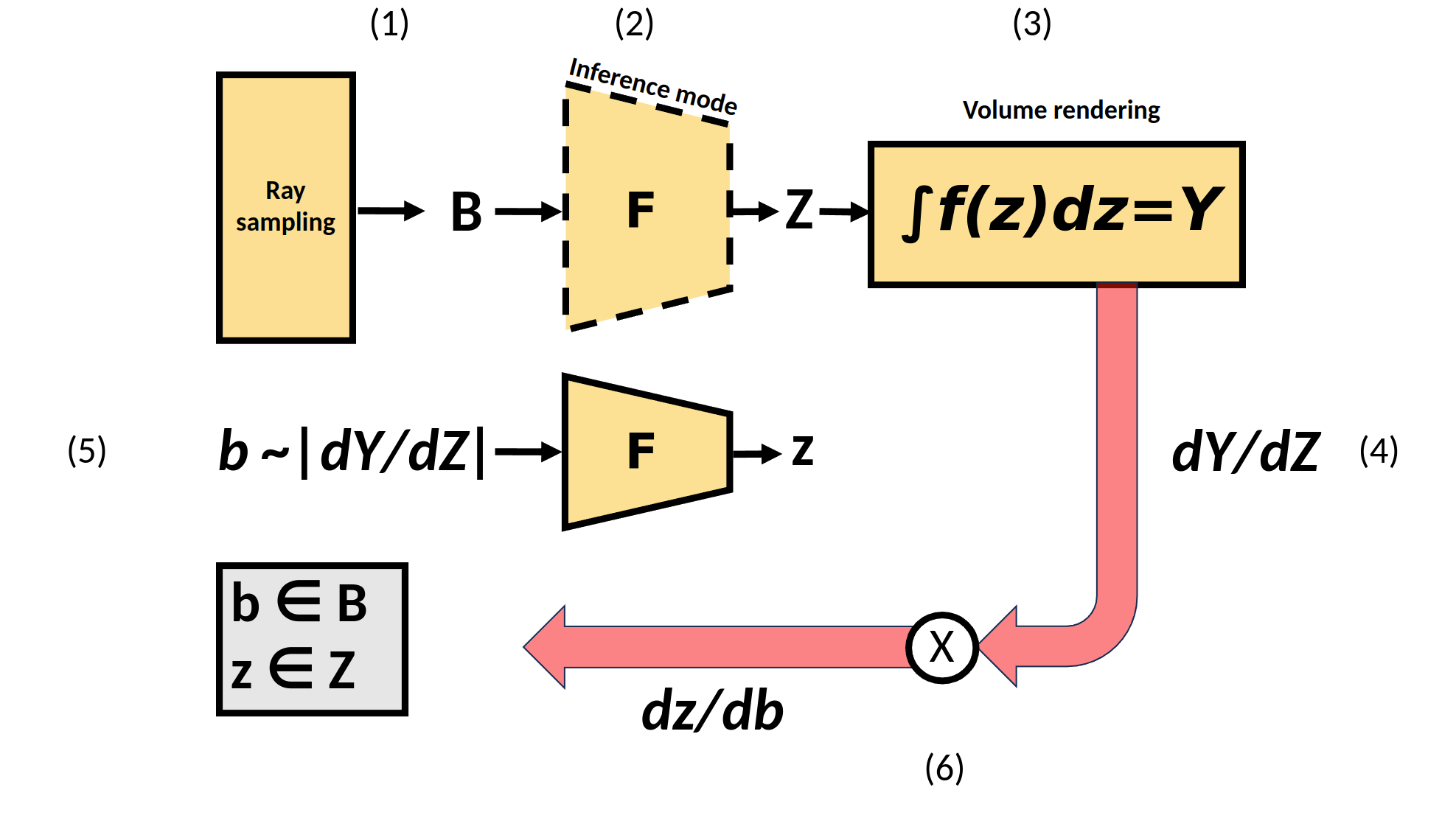}
		\caption{Overview of our hard sample mining process: (1) Cast rays and ray sample producing $B$ ray samples. (2) Run the network in \textit{inference mode}. (3) Volume render and calculate the loss. (4) Backpropagate the loss until pre-activation outputs of the network $Z = (c', \sigma')$ for $B$ samples. (5) Use the norm of the gradient to subsample $B$ for  $b$ hard samples. Rerun the network pass \textit{now building the computational graph}. (6) Continue the backpropagation through the extended computational graph for the $b$ hard samples.}
		\label{fig:overview}
	\end{figure}

	\cref{fig:overview} provides an overview of the proposed hard sample mining method. Here, we discuss the fundamentals of our method and summarize it in \cref{pseudo}.
	
	\subsubsection{Backpropagation as the computational bottleneck} 
	
	Our hard sample mining further identifies the network backward pass as the computational bottleneck when optimizing a NeRF: For a standard deep learning module, such as the multilayer perceptrons used with NeRF, a backward pass has nearly twice the number of operations compared to a forward pass. Additionally, in order to perform the backward pass, it is essential to store the intermediate output tensors of consecutive modules during the forward pass, which consequently results in increased memory use.
	
	\subsubsection{Two-forward-one-backward-pass approach}
	
	Backpropagating a masked loss for a subset of $b$ samples instead of the entire batch of $B$ samples does not yield computational benefits on standard deep learning platforms like PyTorch and Tensorflow. These platforms end up backpropagating zero gradients for the masked-out samples, resulting in no updates while retaining roughly the same compute. We would also still have to save the intermediate tensors for all the samples, since it is impossible to know exactly before-hand which of them prove to be hard samples. Taking inspiration from \cite{shrivastava2016training}, we propose a two-forward-one-backward pass approach for NeRF network processing:
	
	\begin{enumerate}
		\item Perform the first forward pass for the $B$ samples in inference mode (e.g., using \texttt{no\_grad()} in PyTorch). This substantially reduces memory requirements by avoiding the storage of intermediate tensors needed for backward calculations.
		\item Derive the importance sampling distribution and draw the $b$ samples.
		\item Re-run the forward pass for the $b$ samples, building the computational graph.
		\item Perform the backward pass for the $b$ samples.
	\end{enumerate}
	
	This approach takes advantage of our assumption that the ratio of $\frac{b}{B}$ is expected to be small, i.e. we expect only a fraction of the samples to be informative for the update of the NeRF model.
	
	\subsubsection{Propagated pixel loss as the importance sampling distribution}
	\label{distribution}
	
	When sampling for the hard sample minibatch, we want to pick the samples that actually induce a change in the model parameters. This formally translates into a reduced variance of the gradient (VoG) estimates calculated from different instances of the minibatch according to \cite{katharopoulos2019samples}. Optimal sampling distribution to reduce the VoG has been shown to be the full per-sample gradient norm \cite{needell2015stochastic, zhao2015stochastic, alain2016variance}, but calculating it would be computationally too expensive requiring a backward pass per sample. We use the derivations of \cite{katharopoulos2019samples} that the variation of the gradient norm is mostly captured by the gradient of the loss with respect to the pre-activation outputs of the last layer of the neural network.
	
	In the NeRF optimization, the loss is not defined per point sample, but rather per pixel. So to adapt, we backpropagate the pixel loss $L$ over the volume rendering until the pre-activation outputs of density and color MLPs $(c', \sigma')$ and derive the importance sampling distribution as:
	
	\begin{equation}
		G = \|\frac{\partial L}{\partial (c',\sigma')}\|_2.
	\end{equation}
	
	\subsubsection{Dynamic size of the hard sample minibatch}
	\label{dynamic}
	
	We propose to dynamically set the hard sample minibatch size during the training. As the space becomes progressively, adequately modeled during the training, it is reasonable to think that fewer of the stochastically sampled points benefit the model update. 
	
	Let $D$ represents the whole data, $\mathbf{b} \sim G_D$ represents a minibatch (of size $b$) sampled from the data importance sampling distribution, $\mathbf{b_0} \sim U_D$ represents a minibatch (similarly of size $b$) sampled uniformly from the data. However, sampling $\mathbf{b} \sim G_D$ is computationally prohibitive as it would require a forward pass for the whole data at every iteration since $G_D$ evolves due to model updates. Instead lets define $\mathbf{B} \sim U_D$ as a large minibatch sampled uniformly from the data to be a sufficient representative of the whole data. Then we redefine $\mathbf{b} \sim G_B$ and $\mathbf{b_0} \sim U_B$. Using the derivations of \cite{katharopoulos2019samples}, we approximate variance of the gradient reduction when using the minibatch $\mathbf{b}$ instead of $\mathbf{b_0}$, i.e. using importance sampling over uniform sampling, as:
	
	\begin{equation}
		\label{vog}
		R = \frac{1}{\sum_{i=1}^B G_i^2} \|G - U\|_2^2
	\end{equation}
	
	where $G_i = \|\frac{\partial L_i}{\partial (c_i',\sigma_i')}\|_2$ and $U = \frac{1}{B}$ (uniform distribution).
	
	Another commonly accepted proposition in the field of deep learning is that the variance of the gradient is a decreasing polynomial of the minibatch size \cite{qian2020impacttovog}. We assume that if we increase minibatch size by a factor of $\tau$, the variance of the gradient reduces to $1 - \frac{1}{\tau^2}$. Then equating with \cref{vog}, we can solve for $\tau$ as:
	
	\begin{equation}
		\tau = (1 - R)^{-\frac{1}{2}}
		\label{tau}
	\end{equation}
	
	This essentially tells us how much larger should a uniformly sampled minibatch be to achieve the same variance of gradient as the importance sampled minibatch. Let us then define that $\tau b = B$. We then solve for the hard minibatch size $b$ to achieve on average the same variance of the gradient as the large uniform batch of size $B$ as:
	
	\begin{equation}
		\label{minibatch-size}
		b = \frac{B}{\tau}
	\end{equation}
	
	To mitigate the issue that we estimate $\tau$ based on the large minibatch $\mathbf{B}$ instead of the whole data, we keep track of a running average $\hat{\tau} = (1 - \alpha_{\tau}) \hat{\tau} +  \alpha_{\tau} \tau$, updated at every iteration, and use $\hat{\tau}$ instead of $\tau$ in \cref{minibatch-size}. In the experiments we set $\alpha_{\tau}$ = $\frac{1}{\text{num. of training images}}$. We also assume the batch size increase of $\tau$ to reduce the variance by $1 - \frac{1}{\tau^2}$; A reduction of $1 - \frac{1}{\tau}$ is assumed in \cite{qian2020impacttovog} for a shallow neural network, which would lead to even smaller $b$. We choose to remain conservative for the batch size reduction for two reasons: (1) our estimation of $\tau$ comes from minibatch $\mathbf{B}$ rather than the whole data, (2) GPU utilization starts to drop if $b$ shrinks excessively small, and we gain no computational time benefit over the pruned samples. Also, as we discover in the experiments, the first forward pass starts to dictate the max memory requirements after $\frac{b}{B}$ reaches a certain point, so we do not gain additional memory benefits either for additional $b$ shrinkage.

	\subsubsection{Algorithmic implementation}
	\label{algorithm}
	
	Implementation of our method that carries out the hard sample mining for NeRF optimization at a point sample level is given in \cref{pseudo}: Rays and corresponding ground truth colors are picked from the dataset in a standard random way (we note that the existing advanced pixel sampling methods could also be used here). Ray sampling is then carried out to produce a set of $B$ sample points along the rays. First network forward pass of these point samples is carried out in inference mode, i.e. not building the computational graph yet. The computational graph building is switched on for the volume rendering phase and the gradient of the pixel loss is calculated with respect to the pre-activation network outputs (i.e. \textit{attributes'}). The $L_2$ norm of this gradient then forms the importance sampling distribution $G$.
	
	We determine the hard minibatch size $b$ according to \cref{dynamic}. Continuing we draw the $b$ hard points from $G$, and repeat the network forward pass, now building the computational graph for the whole network. We can then continue the backpropagation of the cached gradients through the extended computational graph for the hard point samples.

	\begin{algorithm}[H]
		
		\caption{Hard Sample Mining for NeRF Optimization}\label{pseudo}
		
		\begin{algorithmic}[1]
			\State $\hat{\tau} \gets 1.0$
			\State $\alpha_\tau \gets \frac{1}{\text{len}(training\_images)}$
			\For{each \textit{iteration} in \textit{training\_iterations}}
			\State $rays, gt\_color \gets \text{pixelsampling(}training\_images)$ 
			\State $points \gets \text{raysampling}(rays)$ 
			\State $B \gets \text{len}(points)$ 
			\State \textbf{with no\_grad():} 
			\State \text{\quad} $attributes' \gets \text{nerf}(points, \text{preactivation=True})$
			\State $attributes'.\text{require\_grad}()$
			\State $attributes \gets \text{activation}(attributes')$
			\State $pred\_color \gets \text{volumerendering}(attributes)$
			\State $pixel\_loss \gets \text{loss}(rendered\_color, gt\_color)$
			\State $grad\_attributes \gets \text{autograd}(pixel\_loss, (c', \sigma'))$
			\State $G \gets grad\_attributes.\text{norm}()$
			\State $U \gets 1/B$
			\State $\tau \gets (1 - \frac{1}{ \sum_{i=1}^B G_i^2} ||G - U||_2^2)^{-\frac{1}{2}}$ 
			\State $\hat{\tau} \gets (1 - \alpha_{\tau}) \hat{\tau} +  \alpha_{\tau} \tau$
			\State $b \gets \frac{B}{\hat{\tau}} $ 
			\State $hard\_indices \gets \text{multinomial}(G, b)$
			\State $hard\_points \gets points[hard\_indices]$
			\State $attributes' \gets \text{nerf}(hard\_points, \text{preactivation=True})$ 
			\State $grad\_output \gets \text{cat}(grad\_attributes[hard\_indices])$
			\State $attributes'.\text{backward}(grad\_output)$
			\EndFor
			
		\end{algorithmic}

	\end{algorithm}

	\section{Experiments and Results}
	
	We do all the experiments with the PyTorch reimplementation \cite{torch-ngp} of the Instant-NGP \cite{mueller2022instant} for ease of code modification. Our purpose is to showcase the hard sample mining in action, rather than to reproduce the official paper \cite{mueller2022instant} results. There are many details, like the ray termination and the full-CUDA implementation, that are missing from the PyTorch version which deteriorate the performance in comparison to the official implementation; These improvements are however tangential to the proposed hard sample mining method.
	
	The following analysis of importance sampling distribution is done with the original synthetic NeRF dataset \cite{mildenhall2020nerf}. For the actual evaluation, we use the standard Mip-NeRF-360 dataset \cite{barron2022mipnerf} to offer real-life data performance. We use the evaluation pipeline from \cite{barron2022mipnerf}, except we downscale the outdoor scene images with a factor of 8 (instead of 4 for indoor scenes), bringing all the scene images to a more even standard definition resolution.
	
	From the repository settings, we set the batch size as $B=2^{20}$, and disable iteration based learning rate scheduling since we test against a wall-clock training time. Other options are left as default. The experiments were carried out on a NVIDIA A4500 mobile GPU.
	
	\subsection{Importance sampling distribution analysis}
	
\begin{figure}[tb]
		\centering
		\begin{subfigure}{0.35\linewidth}
			\includegraphics[width=\linewidth]{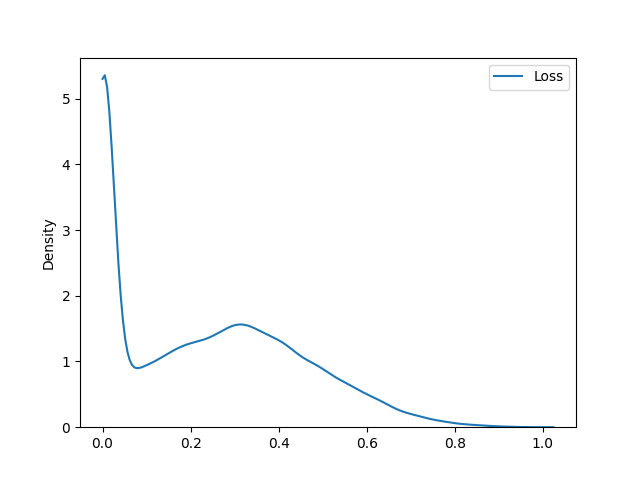}
			\caption{Loss PDF (iter=500)}
			\label{fig:short-a}
		\end{subfigure}
		\hfill
		\begin{subfigure}{0.35\linewidth}
			\includegraphics[width=\linewidth]{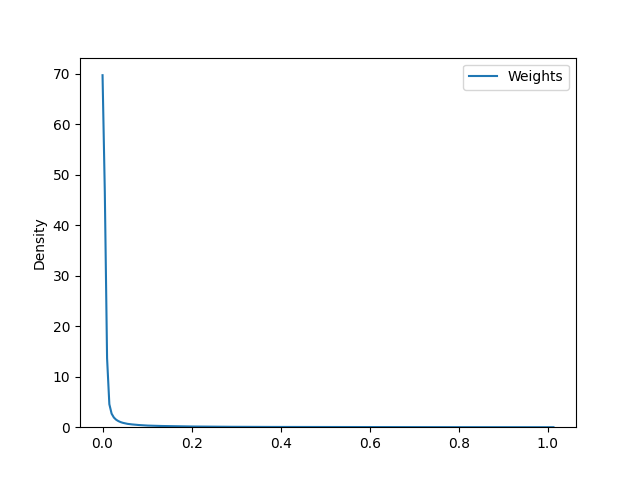}
			\caption{Weight PDF (iter=500)}
			\label{fig:short-a}
		\end{subfigure}
		\hfill
		
		\begin{subfigure}{0.8\linewidth}
			\includegraphics[width=\linewidth]{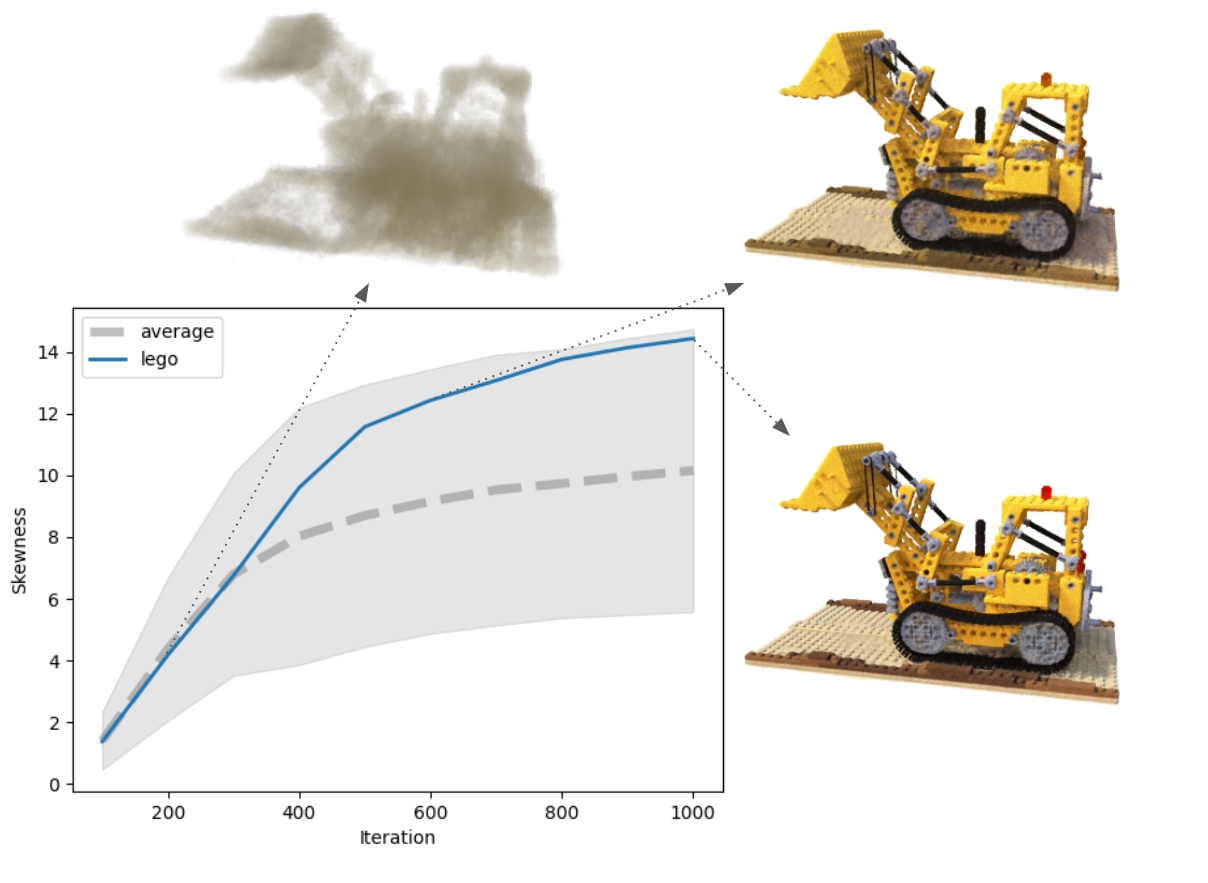}
			\caption{Sample gradient PDF skewness over iterations}
			\label{fig:short-b}
		\end{subfigure}

		\caption{}
		\label{fig:variance}
	\end{figure}

	We start off with a analysis of the importance sampling distribution $G$. We train the baseline Instant-NGP \cite{torch-ngp} for the NeRF synthetic dataset \cite{mildenhall2020nerf}, and calculate the $G$ at every iteration. We form a probability density function (PDF) from the $G$ and plot its skewness value over the whole training in \cref{fig:variance}. The results show that the PDF of the importance sampling distribution becomes increasingly right-tailed. This indicates that there exists variance in the sample importance that our method can take advantage of.
	
	\subsubsection{Inter-ray discrepancy: low loss samples}
	
	We can intuitively understand one major reason why the PDF is skewed: samples belonging to rays that traverse properly modeled regions get backpropagated a low pixel loss, and consequently induce a small gradient themselves. Conversely, samples belonging to high pixel loss rays are exposed to have high-gradients. 
	
	Pixel loss PDF for midpoint iteration is given in \cref{fig:variance}. The pixel loss variance is also exemplified by the lego truck scene renderings at different points of training, where we see the model first converging on major structures of the truck but still struggling with details like the tires, and the individual knobs of the platform. 
	
	\subsubsection{Intra-ray discrepancy: low weight samples.} Low weight samples form another set of samples that are by definition exposed to low gradients. Low weight samples have low impact to the rendered color and subsequently induce only a small error gradient. Sample weight PDF for midpoint iteration is given in \cref{fig:variance}. 
	
\begin{figure*}[tb]
		\centering
		\includegraphics[width=0.5\linewidth]{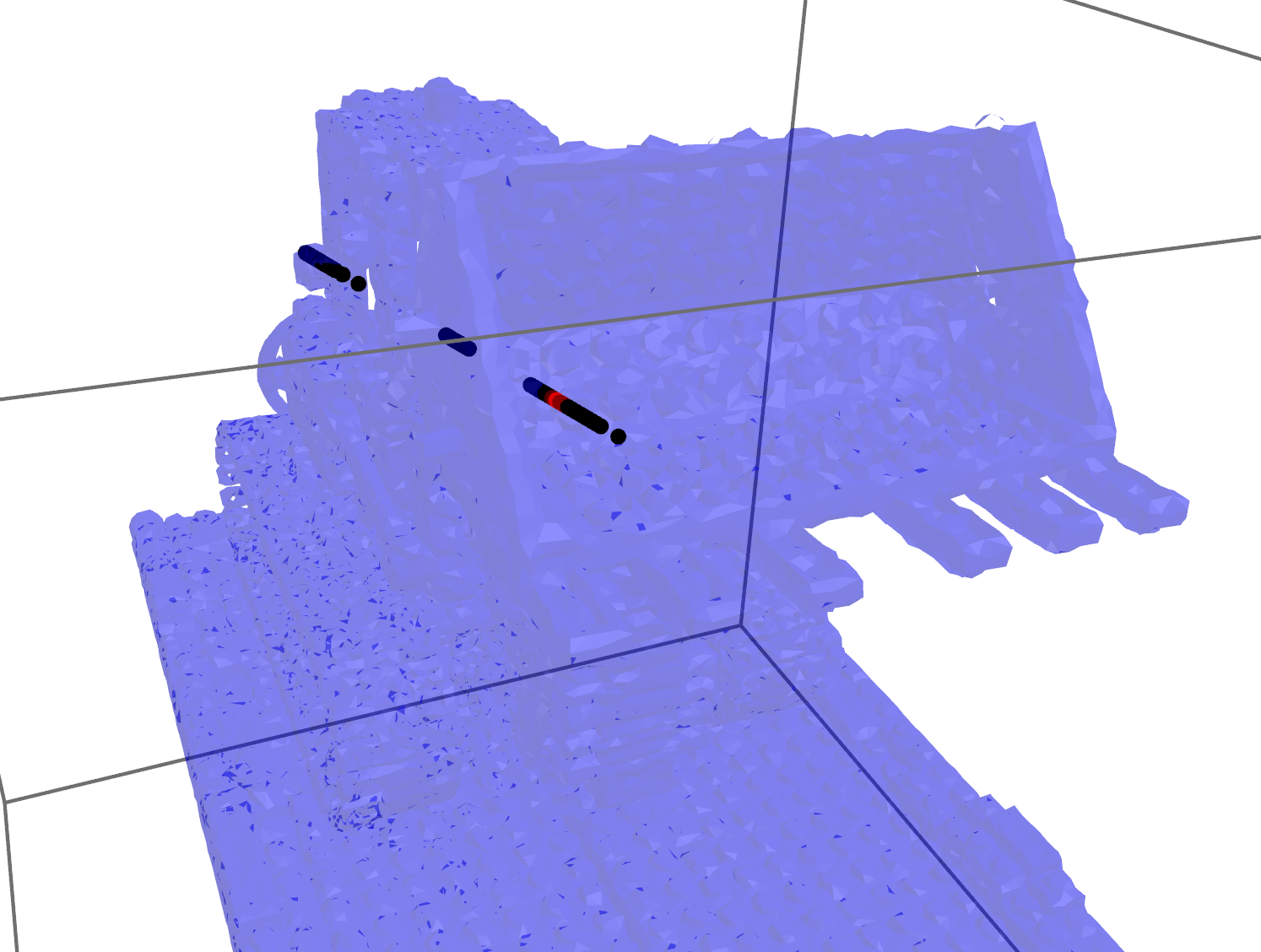}
		\caption{We plot sampled points of one ray during the training. The points are colored
			increasingly red based on their importance sampling weight. For
			visual cue of the scene contents, we also plot the converged mesh model of the lego
			truck into the scene. We can see that the used importance sampling distinguishes the first surface
			colliding samples from empty space (before the collision) and occluded samples (after
			collision) automatically. This is because by definition these latter samples induce near zero gradient.}
		\label{fig:pruning}
	\end{figure*}

	As depicted in \cref{fig:pruning}, we see that the coloring of the sample points based on the sample importance automatically distinguishes the samples at the first surface collision site from samples at empty space before and at occluded space after. These low-weight samples may still exist in the batch after the grid sampling due to the limited occupancy- and hash-grid resolutions, and occluded samples not being pruned by the default method. The results indicate that our method is able to automatically correct for the pruning mistakes of the guided ray sampling algorithms.

	\subsection{Runtime and Memory Reduction}
	
	\textbf{Runtime.} The network forward and backward pass on hard samples, takes roughly $\frac{b}{B}$ of the original wall-clock runtime. Combined with the $\frac{1}{3}$ coming from the first forward pass in inference mode, we achieve a runtime reduction of $\frac{B+3b}{3B}$ for the network processing. We can use this as a low-end approximation of the whole iteration time reduction, that does not take into account the time taken by the ray sampling and volume rendering modules (which remain to have the original runtime).
	
	In the experiments, as depicted in \cref{tab:iNGP-360}, we see a reduction of the average iteration time down to $\sim$53-54\% when our hard sample batch fraction $\frac{b}{B}$ is on average $\sim$10-15\%. The roughly $\sim$10\% above the low-end approximation is contributed to the time taken by the ray sampling and volume rendering modules.
	
	\textbf{Memory usage.} The first network forward pass is run in evaluation mode and no intermediate tensors are required to be saved for backward pass yet. Since the computational graph is only built during the second forward pass with the reduced batch size $b$ we see a drastic reduction in GPU memory usage.
	
	Memory usage per iteration is plotted in \cref{fig:performance} averaged over the Mip-NeRF-360 dataset scenes for both the baseline and our hard sample mining. We see common phases in both memory curves: (1) Peak memory usage happens due to occupancy grid update with large sample size during the first 256 iterations. (2) Warmup iterations where adaptive number of rays~\cite{mueller2022instant} and/or $\frac{b}{B}$ ratio is stabilizing. (3) Network memory usage and upticks from continuing to update the occupancy grid.
	
	We see that the essential statistic for other applications is the reduction in the network memory usage from the baseline's $\sim$1 GB to $\sim$570 MB by our hard sample mining. The regular upticks in memory usage are caused by the occupancy grid updates, which is out the scope of our method.
	
	\begin{table}[tb]
		\footnotesize
		\centering
		\caption{Performance statistics of applying hard sampling to training the Instant-NGP for the Mip-NeRF-360-dataset scenes. Average PSNR over all the validation images are reported for 2, 5, and 8 minutes of training. $\tau^{-1}$ describes the converged $\frac{b}{B}$.  $\frac{ms}{iter}$ is the average training iteration time in milliseconds per iteration. We report in format \textit{hard sample mining \,$|$\, baseline} when comparing our hard sample mining to baseline training performance. \textbf{*} denotes the repository reported benchmark results for 8 minutes of training the baseline.}
		\label{tab:iNGP-360}
		\begin{tabularx}{\textwidth}{X *{7}{l}} 
			\toprule
			& bonsai & kitchen & room & counter & garden & bicycle & stump \\
			\midrule
			2\,min & 27.9\,$|$\,26.8 & 25.7\,$|$\,24.1 & 27.9\,$|$\,27.3 & 24.7\,$|$\,23.8 & 25.4\,$|$\,24.5 & 21.8\,$|$\,21.3 & 23.0\,$|$\,22.8 \\
			5\,min & 29.2\,$|$\,28.1 & 26.9\,$|$\,26.0 & 28.8\,$|$\,28.3 & 25.4\,$|$\,24.9 & 26.5\,$|$\,25.8 & 22.2\,$|$\,22.1 & 23.3\,$|$\,23.3 \\
			8\,min & \textbf{29.7}\,$|$\,28.4 & \textbf{27.5}\,$|$\,26.7 & \textbf{29.1}\,$|$\,28.7 & \textbf{25.6}\,$|$\,25.3 & \textbf{26.9}\,$|$\,26.3 & \textbf{22.4}\,$|$\,22.1 & \textbf{23.4}\,$|$\,23.4 \\
			
			8\,min* & 29.0 & 26.4 & 28.6 & 25.2 & 23.7 & 21.3 & 22.7 \\
			\midrule
			\(\tau^{-1}\) & 0.10 & 0.11 & 0.13 & 0.11 & 0.14 & 0.11 & 0.15 \\
			\midrule
			$\frac{ms}{iter}$ & 31\,\,$|$\,\,58 & 35\,\,$|$\,\,64 & 33\,\,$|$\,\,61 & 34\,\,$|$\,\,64 & 37\,\,$|$\,\,60 & 32\,\,$|$\,\,56 & 37\,\,$|$\,\,68 \\
			
			\bottomrule
		\end{tabularx}
	\end{table}
	\subsection{View-Synthesis}
	
	\subsubsection{Training performance.} In \cref{fig:performance}, we plot the training loss per iteration averaged across all the 7 scenes in the Mip-NeRF-360 dataset \cite{barron2022mipnerf}. We see our hard sample mining closely following the baseline training loss per iteration. This indicates that training on the samples that our hard sample mining prunes has a negligible effect on the training loss. Due to the significantly reduced training iteration time, our hard sample mining method allows running substantially higher number of iterations and reaches significantly lower loss within the same 8-minute wall-clock training time period.
	
	\subsubsection{Validation performance} 
\begin{figure}[tb]
		\centering
		\begin{subfigure}{0.32\linewidth}
			\includegraphics[width=\linewidth]{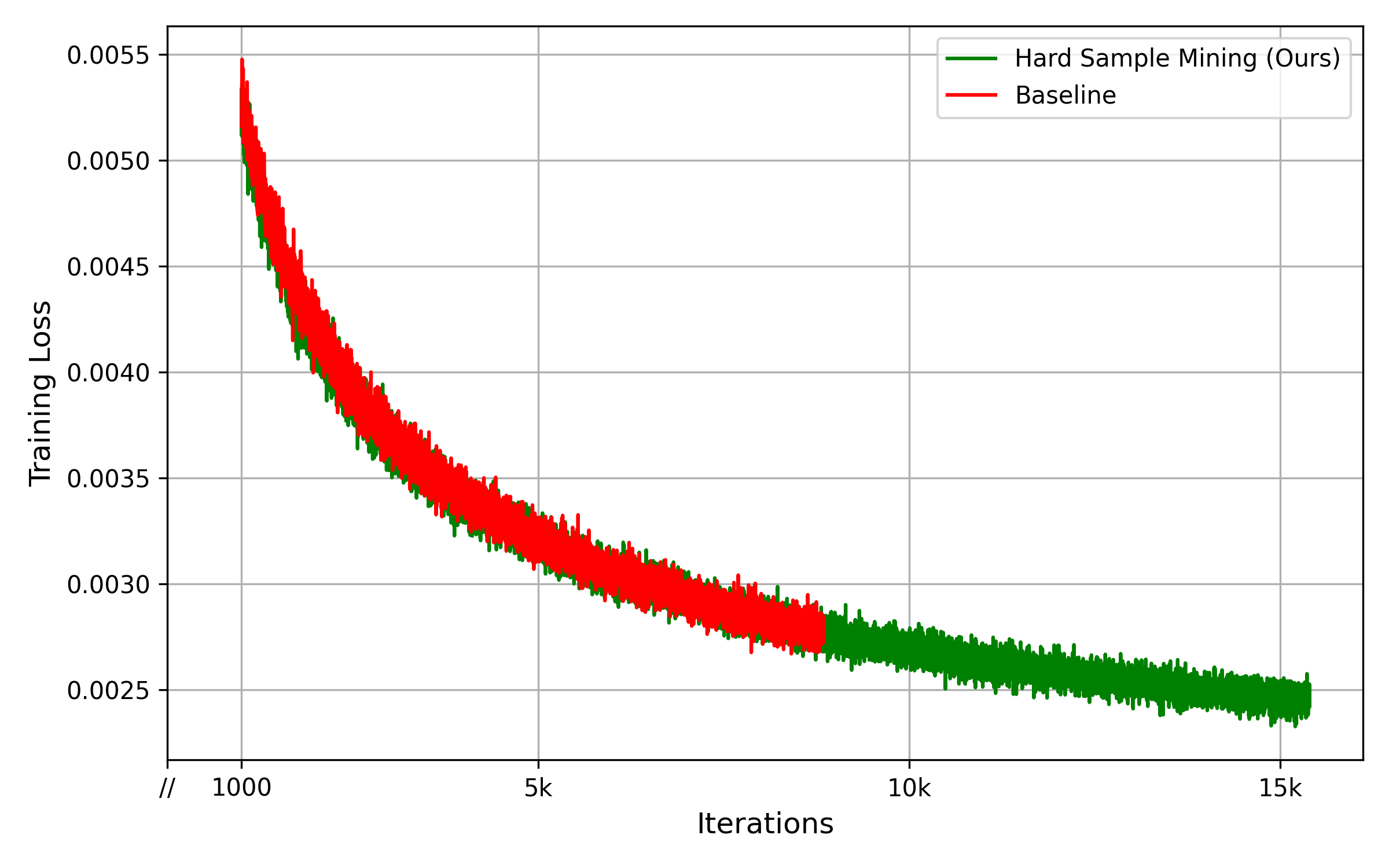}
			\caption{Training loss }
			\label{fig:short-a}
		\end{subfigure}
		\hfill
		\begin{subfigure}{0.32\linewidth}
			\includegraphics[width=\linewidth]{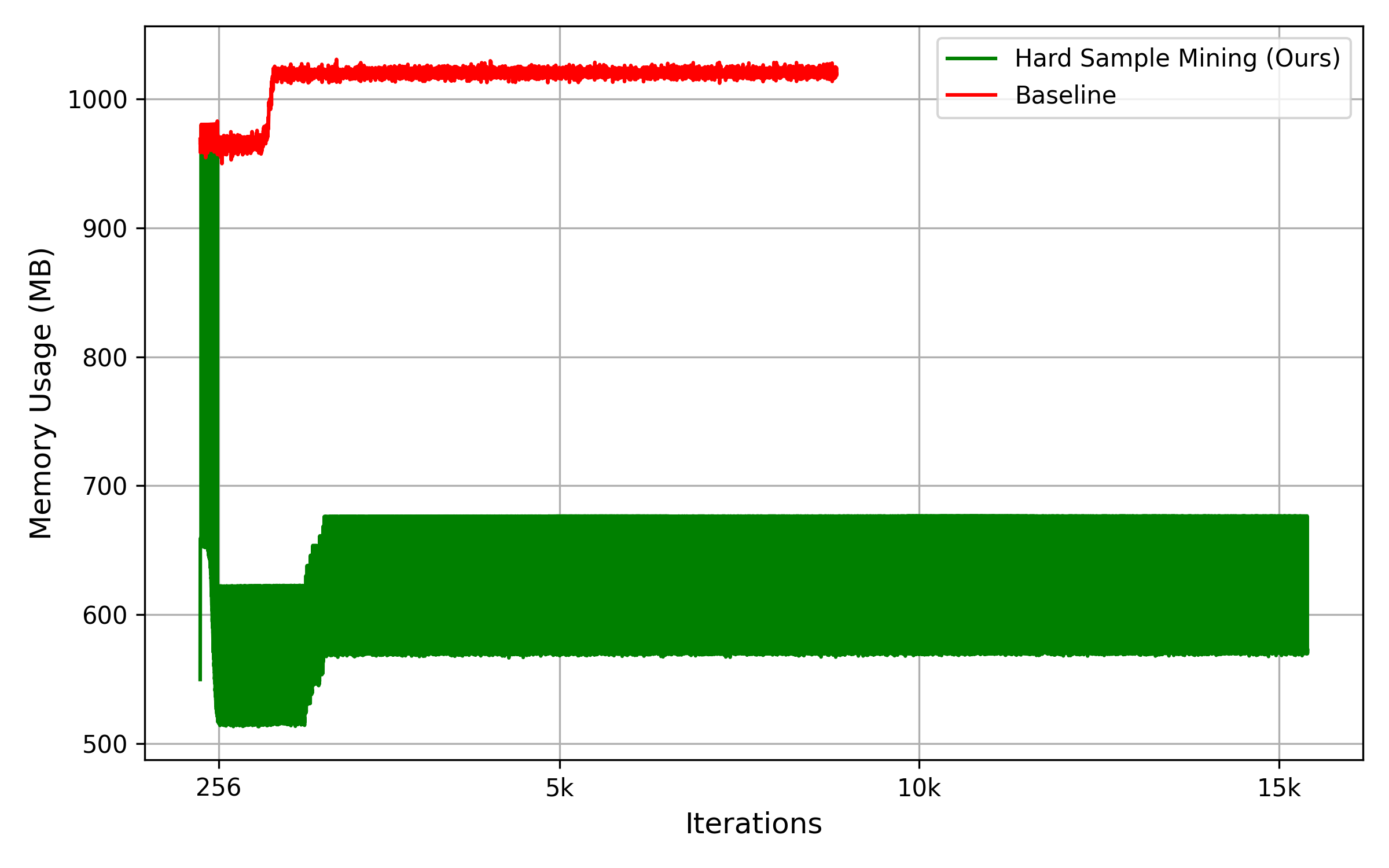}
			\caption{Memory usage }
			\label{fig:short-b}
		\end{subfigure}
		\hfill
		\begin{subfigure}{0.32\linewidth}
			\includegraphics[width=\linewidth]{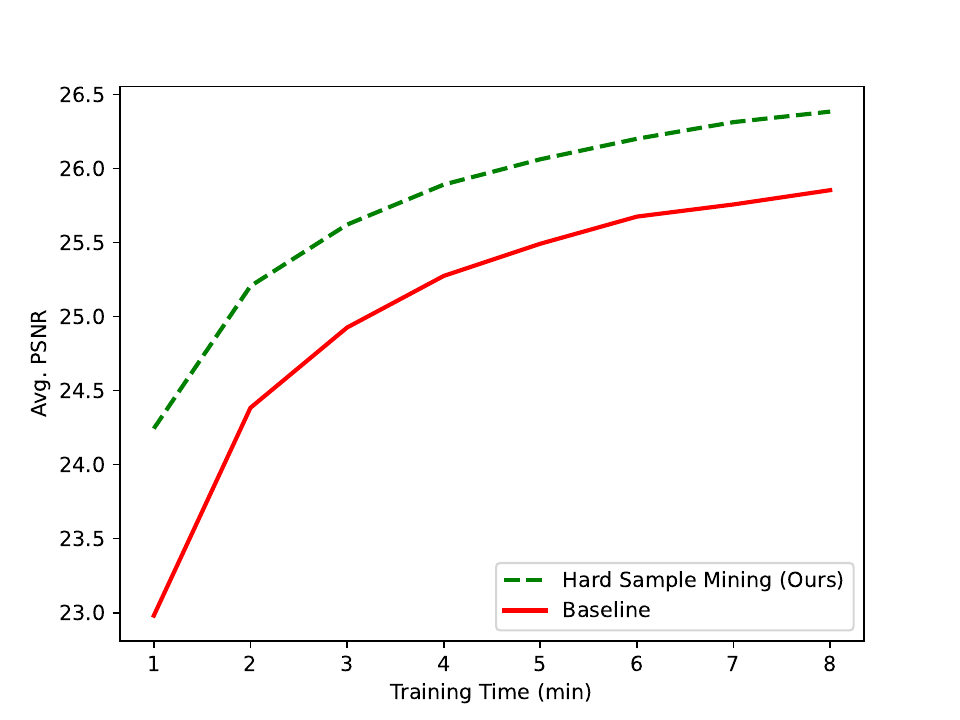}
			\caption{Validation PSNR }
			\label{fig:short-b}
		\end{subfigure}
		\caption{Performance over training time: (a) We see our hard sample mining closely following the training loss of the baseline \textit{per iteration}, and reducing the (b) training time memory usage. Due to reducing iteration time, our hard sample mining gets to run for more iterations, and achieves higher (c) validation PSNR per training time. All statistics are averaged over the Mip-NeRF-360 scenes.}
		\label{fig:performance}
	\end{figure}

For quantitative results, \cref{fig:performance} indicates average validation PSNR performance over all the Mip-NeRF-360 scenes. We can see our hard sample mining beating the baseline convincingly over the whole training time. It converges much faster ($\sim$2x) to a given PSNR level, and has $\sim$1 dB gain per same training time.
	
\begin{figure}[tb]
		\centering
		\begin{tikzpicture}
			\draw (0, 0) node[inner sep=0] {\includegraphics[width=\linewidth]{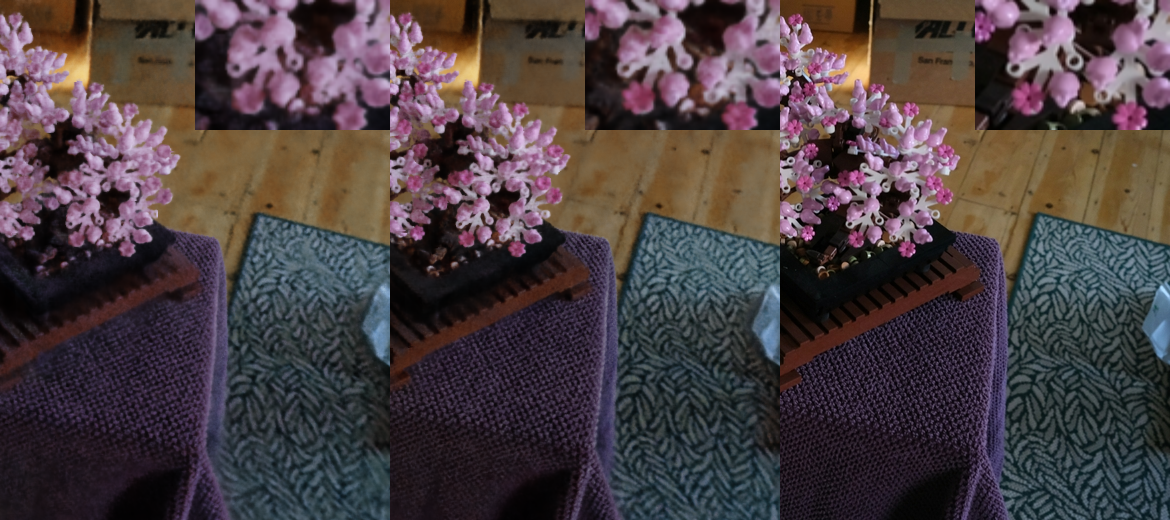}};
			\draw (-4.5, -2.5) node[text=white] {Baseline/PSNR 28.38};
			\draw (-0.2, -2.5) node[text=white] {HSM(Ours)/PSNR 29.38};
		\end{tikzpicture}
		\resizebox{\linewidth}{!}{%
			\begin{tikzpicture}
				\draw (0, 0) node[inner sep=0] {\includegraphics[width=\linewidth]{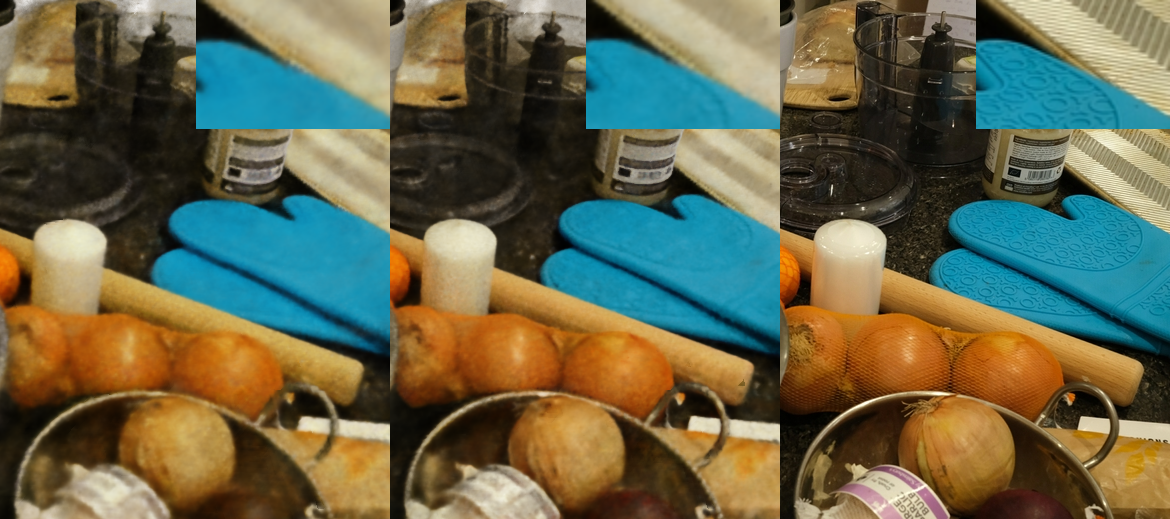}};
				\draw (-4.5, -2.5) node[text=white] {Baseline/PSNR 21.87};
				\draw (-0.2, -2.5) node[text=white] {HSM(Ours)/PSNR 22.31};
			\end{tikzpicture}
			\begin{tikzpicture}
				\draw (0, 0) node[inner sep=0] {\includegraphics[width=\linewidth]{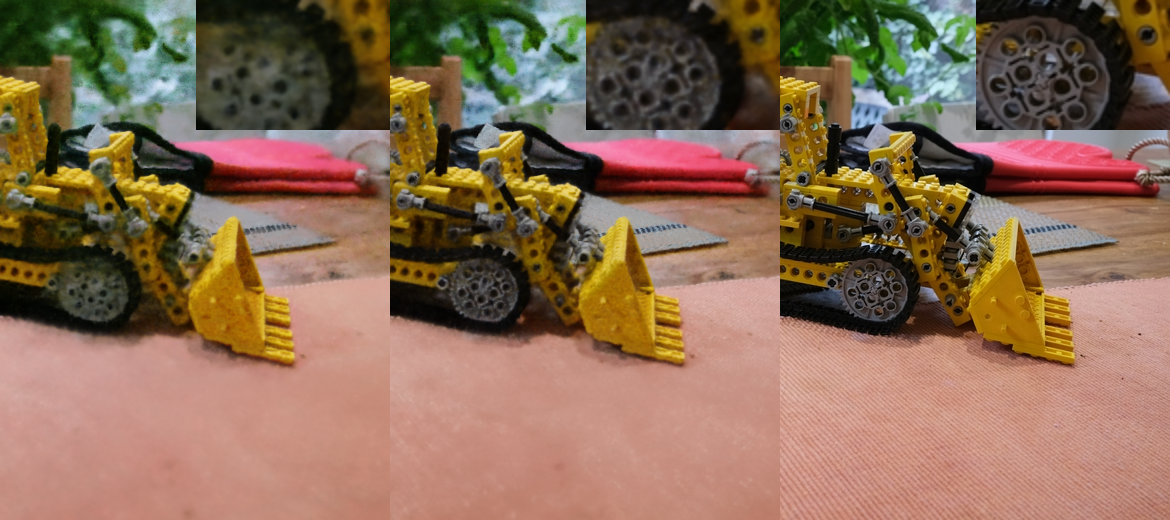}};
				\draw (-4.5, -2.5) node[text=white] {Baseline/PSNR 22.32};
				\draw (-0.2, -2.5) node[text=white] {HSM(Ours)/PSNR 23.93};
			\end{tikzpicture}
			
		}
		\resizebox{\linewidth}{!}{%
			\begin{tikzpicture}
				\draw (0, 0) node[inner sep=0] {\includegraphics[width=\linewidth]{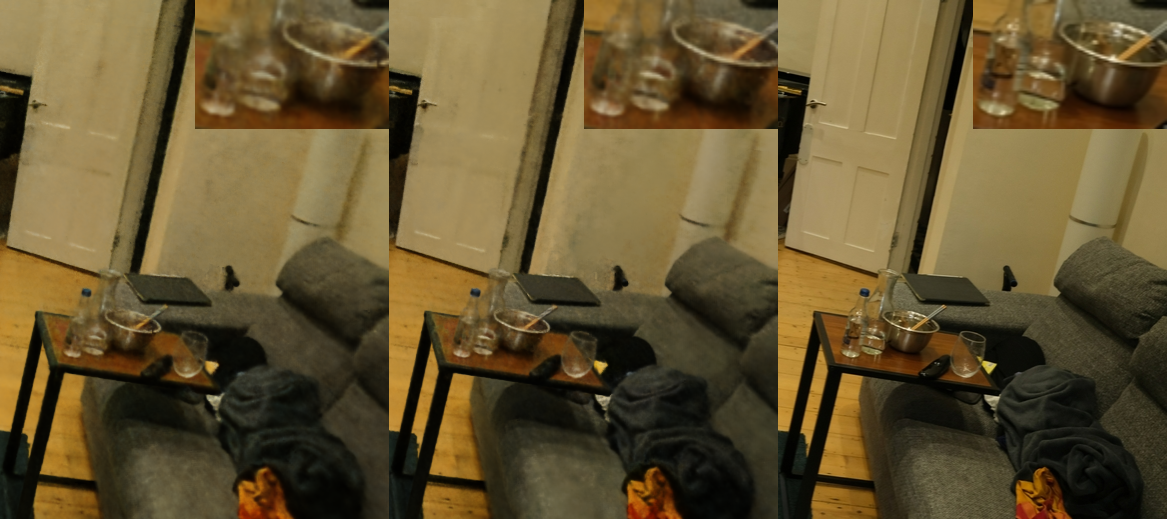}};
				\draw (-4.5, -2.5) node[text=white] {Baseline/PSNR 29.74};
				\draw (-0.2, -2.5) node[text=white] {HSM(Ours)/PSNR 30.63};
			\end{tikzpicture}
			
			\begin{tikzpicture}
				\draw (0, 0) node[inner sep=0] {\includegraphics[width=\linewidth]{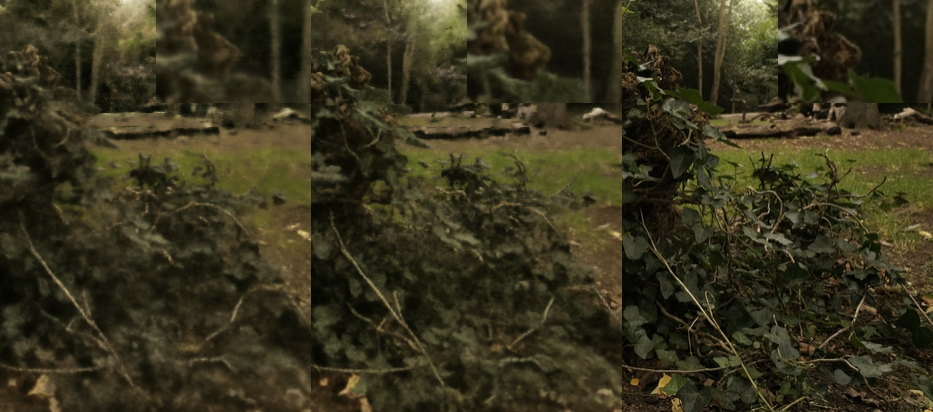}};
				\draw (-4.5, -2.5) node[text=white] {Baseline/PSNR 23.73};
				\draw (-0.2, -2.5) node[text=white] {HSM(Ours)/PSNR 24.43};
			\end{tikzpicture}
		}
		\resizebox{\linewidth}{!}{%
			\begin{tikzpicture}
				\draw (0, 0) node[inner sep=0] {\includegraphics[width=\linewidth]{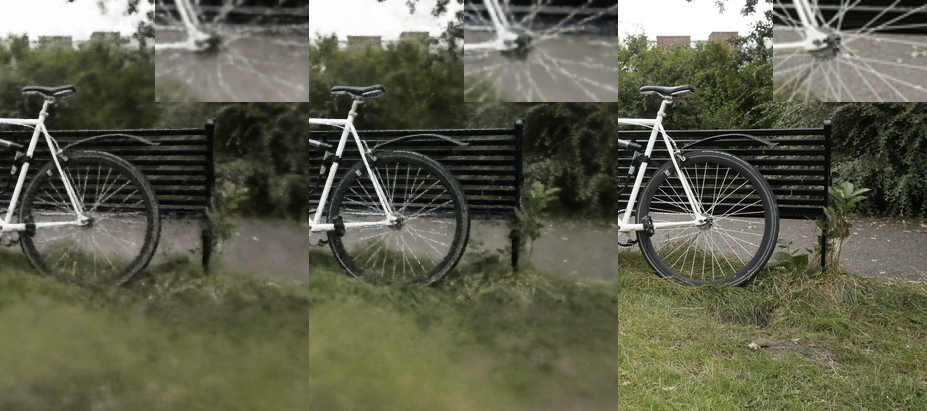}};
				\draw (-4.5, -2.5) node[text=white] {Baseline/PSNR 20.06};
				\draw (-0.2, -2.5) node[text=white] {HSM(Ours)/PSNR 20.46};
			\end{tikzpicture}
			\begin{tikzpicture}
				\draw (0, 0) node[inner sep=0] {\includegraphics[width=\linewidth]{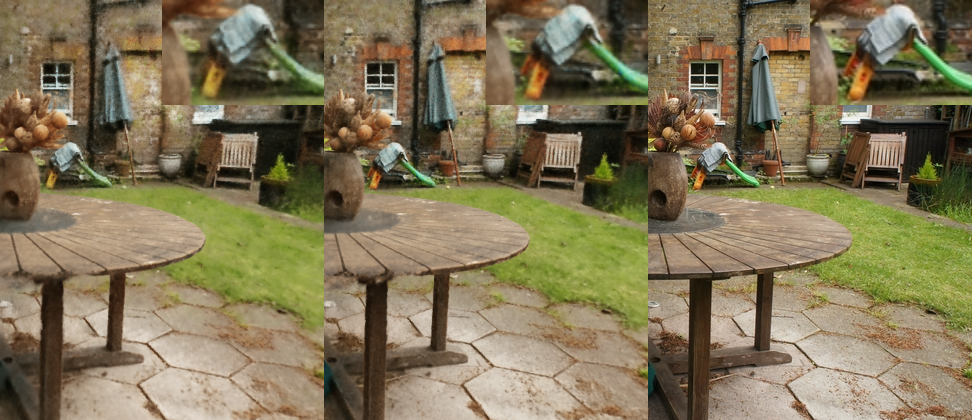}};
				\draw (-4.5, -2.5) node[text=white] {Baseline/PSNR 25.35};
				\draw (-0.2, -2.5) node[text=white] {HSM(Ours)/PSNR 26.27};
			\end{tikzpicture}
		} 
		\caption{Rendered validation images after training an Instant-NGP model for 2 minutes for the scenes of the Mip-NeRF-360 dataset. We see our hard sample mining (HSM) method achieving better quality especially on details compared to the baseline.}
		\label{fig:renders-indoors}
	\end{figure}

	For qualitative results, \cref{fig:renders-indoors} shows validation image renderings after just 2 minutes of training. We see our hard sample mining achieving a better overall appearance, and especially better details: as a representative example at the uppermost row we see that our hard sample mining training has already achieved to model the fine structures of the lego bonsai tree when the random sampling still struggles to differentiate between the leaves and the branches. 
	
	\subsubsection{Variability in hard sample mining effectiveness}. We observe that a indicator of the effectiveness of our hard sample mining method is $\tau^{-1}$, which essentially tells us the ratio $\frac{b}{B}$. Subsequently, we see the achieved Peak Signal-to-Noise Ratio (PSNR) gain achieved with our method is inversely aligned with this metric.
	
	Looking at \cref{tab:iNGP-360} for the converged $\tau^{-1}$ values per scene, we notice that scenes like room, garden, and stump have notably higher values. These scenes seem to have more consistent difficulty across the scene, with characteristics such as unboundedness and/or multiple outward-facing objects. On the other hand, scenes with low values (bonsai, kitchen, counter, and bicycle) share a common characteristic - they have centrally placed objects against a relatively easy background. Uniformly challenging content in a scene seems to limit the anticipated computational efficiency gains resulting from our method. We attribute this to the pixel loss distribution of the samples being more uniform. 
	
	\section{Discussion \& Conclusions}
	
	We showcase a hard sample mining algorithm that employs propagated pixel loss to rank the point samples and subsequently subsamples a hard  minibatch, estimated to retain most of the update information. The construction of the network computational graph and parameter updates based on only this hard sample minibatch reduces the iteration time and the memory usage during the training. This optimization strategy applied to the PyTorch reimplementation \cite{torch-ngp} of the Instant-NGP \cite{mueller2022instant} effectively enhances the novel view synthesis quality achieved with the tested Mip-NeRF-360 dataset \cite{barron2022mipnerf} scenes with reduced training time and memory usage. We demonstrated the efficiency improvements of integrating our hard sample mining into neural radiance fields (NeRF) optimization in \cref{fig:performance}. 
	
	Our results indicate that it is beneficial to take the pixel loss into account when doing importance sampling of the point samples. In this regard, our method is first-of-its-kind as far as we know. The provided speed-up to the training by focusing on the hard samples could be very useful for slow methods like \cite{barron2022mipnerf, li2023neuralangelo}. The hyperparameter-free way of automatically determining the hard minibatch size also offers an interesting solution to a relevant problem of the optimal batch size when doing NeRF optimization. It enables the utilization of large batch sizes for sufficient coverage of the scene, without excessive memory usage. We see that this idea could be very useful for the methods \cite{barron2022mipnerf, li2023neuralangelo}, but also to the prominent 3D gaussian splatting approach \cite{3dgs}, which suffers from extremely high memory usage during training.
	
	In the future work, we aim to investigate the interactions of our method with various NeRF techniques with diverse scenes. To achieve this, we plan to implement our method in NeRFStudio~\cite{nerfstudio}, which gives access to multiple NeRF models and datasets.  Implementation in NeRFStudio also allows us to make a comparison to Nerfacc \cite{li2023nerfacc} way of just pruning the low-weight samples. We anticipate our method to be easily integratable as it only interfaces with the NeRF network module. The impact of pose errors in the dataset also remains mostly an unexplored area. It is possible that severe pose mistakes could produce "impossible to learn" point samples that consistently crowd the hard minibatch. To this end, we could apply the available techniques like uncertainty \cite{martinbrualla2021nerf} and pose \cite{wang2022nerf--} estimation in NeRFStudio, to mitigate the possible issues caused by the pose mistakes.

	\clearpage
	
\appendix	
	
\section{Additional baseline method details}
We use the open-sourced PyTorch reimplementation of the Instant-NGP at \href{https://github.com/ashawkey/nerf_template}{\text{https://github.com/ashawkey/nerf\_template}} and adapt the default training script for the Mip-NeRF-360 scenes by increasing the batch size to $B=2^{20}$, disabling the iteration based learning rate (since we test against the wall-clock training time) and downscaling the outdoor images with a factor of 8 (instead of 4 for indoor), to bring both the indoor and the outdoor images to a more even standard definition resolution. We follow the guideline of the \cite{barron2022mipnerf} by using 7/8 images for training and 1/8 images, evenly distributed, for validation.

We note that the PyTorch version does not reimplement every feature of the official implementation~\cite{mueller2022instant}. Most notably the ray termination and the full-CUDA-coding are left out. This leads to the PyTorch implementation not reaching the same performance as the official implementation, which can be seen from the NeRF synthetic dataset results reported in \cref{tab:syn}. We plan on a NeRFStudio implementation for more comprehensive testing.

\section{Additional results}

\subsection{Converged modeling results}

\begin{figure}[tb]
	\centering
	\begin{subfigure}{0.48\linewidth}
		\includegraphics[width=\linewidth]{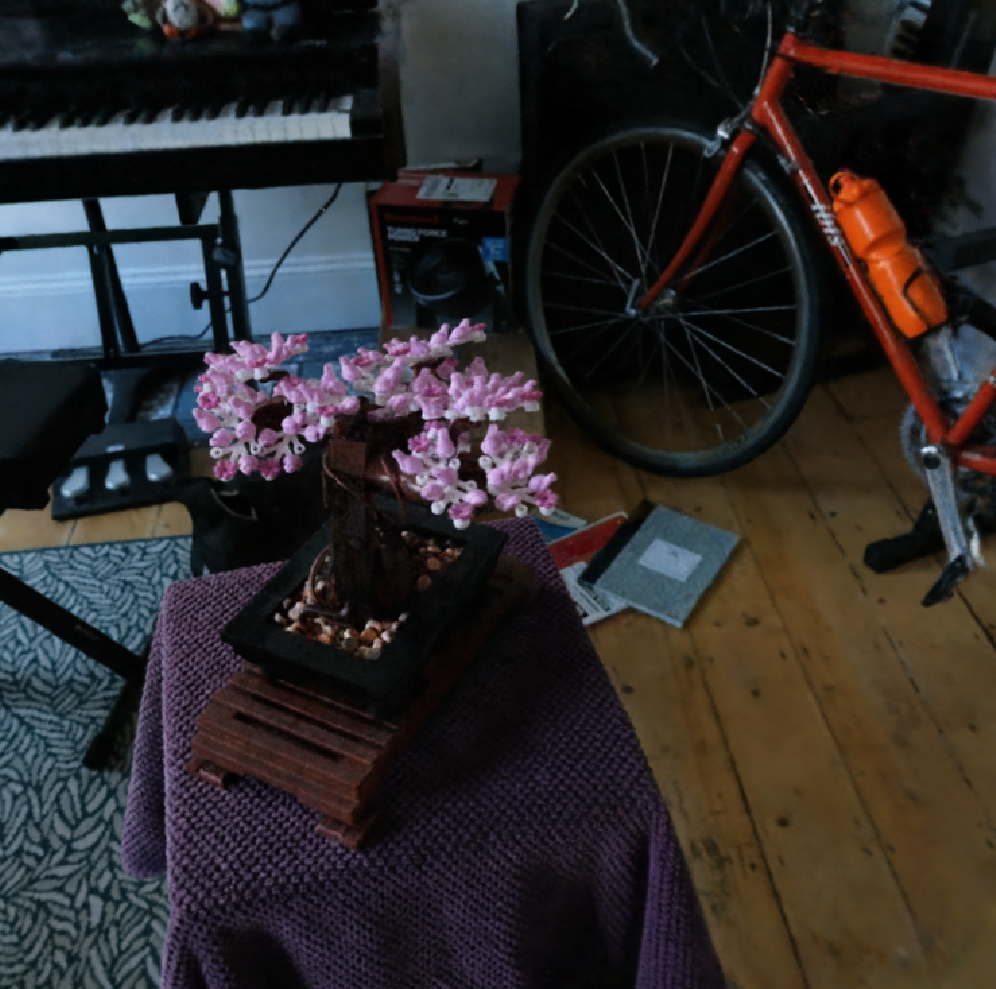}
		\caption{Bonsai}
		\label{fig:short-a}
	\end{subfigure}
	\hfill
	\begin{subfigure}{0.48\linewidth}
		\includegraphics[width=\linewidth]{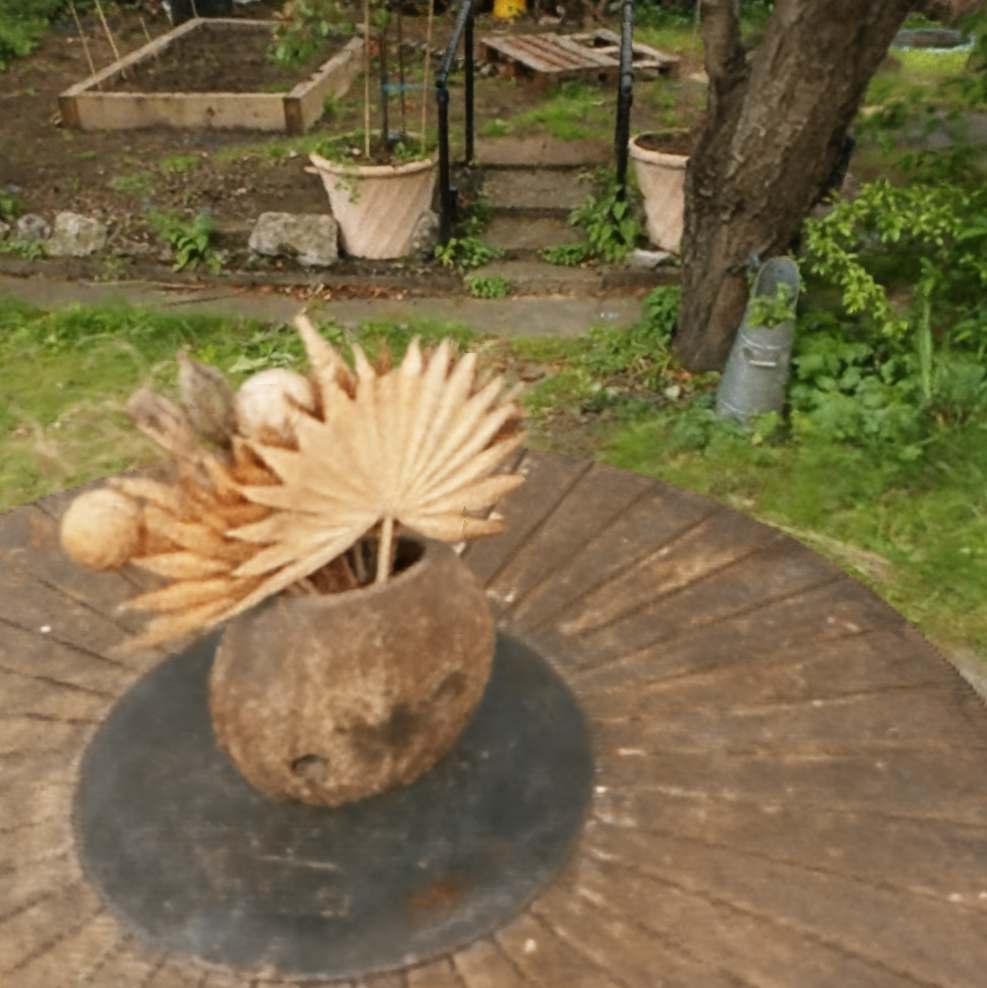}
		\caption{Garden}
		\label{fig:short-b}
	\end{subfigure}
	\caption{Free-viewpoint test time renderings for the Instant-NGP trained with the hard sample mining.}
	\label{fig:freeview}
\end{figure}

We report the converged training results for the Instant-NGP on the Mip-NeRF-360 dataset scenes in \cref{tab:iNGP-360-long}, letting the training continue for 20 minutes. In \cref{fig:freeview}, we visualize free-viewpoint renderings using the repository's GUI renderer; We observe no test time artifacts caused by using our hard sample mining method during the training.

\begin{table}[h]
	\tiny 
	\centering
	\caption{Validation dataset Peak Signal-to-Noise-Ratio when training Instant-NGP for 20 min on the Mip-NeRF-360 dataset. Results reported in format \textit{hard sample mining}~|~\textit{baseline}.}
	\label{tab:iNGP-360-long}
	\begin{tabularx}{\textwidth}{X *{7}{>{\centering\arraybackslash}p{1.35cm}}} 
		\toprule
		& bonsai & kitchen & room & counter & garden & bicycle & stump \\
		\midrule
		20~min & 30.35~|~30.03 & 28.27~|~27.65 & 29.42~|~29.05 & 26.09~|~26.00 & 27.25~|~26.84 & 22.66~|~22.36 & 23.28~|~23.46 \\
		\bottomrule
	\end{tabularx}
\end{table}

\subsection{NeRF synthetic dataset results}

We report the Instant-NGP results for the NeRF synthetic dataset \cite{mildenhall2020nerf} in \cref{tab:syn}. We use the train split for training and evaluate the PSNR on the test split. As batch size we use $B=2^{18}$ as mentioned in \cite{mueller2022instant}.

\begin{table}[h]
	\tiny 
	\centering
	\caption{Test dataset Peak Signal-to-Noise-Ratio when training Instant-NGP on the NeRF synthetic dataset. Results reported in format \textit{hard sample mining}~|~\textit{baseline}. \textbf{*} denotes the official paper \cite{mueller2022instant} results for 5 minutes of training.}
	\label{tab:syn}
	\begin{tabularx}{\textwidth}{X *{8}{>{\centering\arraybackslash}p{1.30cm}}} 
		\toprule
		& mic & ficus & chair & hotdog & materials & drums & ship & lego \\
		\midrule
		1~min & 35.18~|~31.77 & 32.64~|~31.32 & 34.48~|~32.11 & 36.43~|~34.29 & 28.42~|~26.51 & 25.49~|~24.72 & 29.12~|~26.86 & 34.98~|~32.00 \\
		5~min & 35.76~|~34.12 & 32.21~|~32.21 & 34.87~|~33.22 & 36.66~|~35.15 & 28.69~|~27.18 & 25.58~|~25.17 & 29.53~|~27.55 & 35.19~|~32.74 \\
		5~min* & 36.22 & 33.51 & 35.00 & 37.40 & 29.78 & 26.02 & 31.10 & 36.39 \\

		\bottomrule
	\end{tabularx}
\end{table}

\subsection{Nerfacto results}

We also did testing of our method with the Nerfacto \cite{nerfstudio} reimplementation in the repository \cite{torch-ngp}, which uses the proposal sampling method for its ray sampling. We use a lower batch size ($B=2^{18}$ instead of $B=2^{20}$) since with the Nerfacto architecture we need to process on average $5.3$ proposal samples per main NeRF network sample. Also, to make the results comparable to the ones reported in \cite{nerfstudio}, we did not apply any separate downscaling for the outdoor scenes.

The results seem promising for the generalizability of our method: The dynamic memory footprint is reduced from 300 MB to 160 MB for normal iterations when using our hard sample mining (570 MB to 430 MB for iterations when sampler networks are updated). However, the measured PSNR gains per training time are not as large as with the Instant-NGP. We contribute this to two factors: The proposal sample processing, which is out-of-the-scope of our hard sample mining, (1) forms a larger part of the computation, and (2) prunes away the low-weight samples more effectively when compared to the grid sampling of Instant-NGP.

\begin{table}[h]
	\tiny 
	\centering
	\caption{Validation dataset Peak Signal-to-Noise-Ratio when training Nerfacto on the Mip-NeRF-360 dataset. Results reported in format \textit{hard sample mining}~|~\textit{baseline}. We also report the results from the official paper~\cite{nerfstudio} in the second section.}
	\label{tab:iNGP-360}
	\begin{tabularx}{\textwidth}{X *{7}{>{\centering\arraybackslash}p{1.35cm}}} 
		\toprule
		& bonsai & kitchen & room & counter & garden & bicycle & stump \\
		\midrule
		2~min & 27.88~|~27.57 & 27.11~|~26.42 & 29.15~|~28.56 & 24.96~|~24.60 & 23.37~|~23.26 & 22.06~|~21.98 & 23.92~|~23.69 \\
		4~min & 29.75~|~29.46 & 29.44~|~29.01 & 30.58~|~30.18 & 25.95~|~25.86 & 24.44~|~24.30 & 22.96~|~22.96 & 24.83~|~24.69 \\
		8~min & 30.82~|~30.68 & 30.53~|~30.24 & 31.46~|~31.38 & 26.64~|~26.63 & 25.22~|~25.20 & 23.56~|~23.67 & 25.49~|~25.33 \\
		12~min & 31.31~|~31.18 & 30.96~|~30.74 & 31.79~|~31.83 & 26.94~|~26.97 & 25.58~|~25.61 & 23.84~|~23.98 & 25.76~|~25.63 \\
		\midrule
		\midrule
		NeRF & 26.81 & 26.31 & 28.56 & 25.67 & 23.11 & 21.76 & 21.73 \\
		MipNeRF & 27.13 & 26.47 & 28.73 & 25.59 & 23.16 & 21.69 & 23.10 \\
		NeRF++ & 29.15 & 27.80 & 28.87 & 26.38 & 24.32 & 22.64 & 24.34 \\
		MipNeRF-360 & 33.46 & 32.23 & 31.63 & 29.55 & 26.98 & 24.37 & 26.4 \\
		Nerfacto(5min)\cite{nerfstudio} &  28.98 &  28.17 &  29.36 &  25.92 &  24.05 &  22.36 &  18.94 \\
		
		\bottomrule
	\end{tabularx}
\end{table}

\clearpage


%
%
\bibliographystyle{splncs04}
\bibliography{main}
\end{document}